\documentclass{article}

\PassOptionsToPackage{numbers, compress}{natbib}

\usepackage[preprint]{neurips_2026}

\usepackage[utf8]{inputenc}
\usepackage[T1]{fontenc}
\usepackage{hyperref}
\usepackage{url}
\usepackage{booktabs}
\usepackage{amsfonts}
\usepackage{nicefrac}
\usepackage{microtype}
\usepackage{xcolor}
\usepackage{amsmath}
\usepackage{graphicx}

\usepackage{tikz}
\usepackage{pgfplots}
\pgfplotsset{compat=1.18}

\title{STATERA: Hidden Mass Estimation via Zero-Shot Sim-to-Real Kinematics using Frozen Temporal Tubelets}

\author{
  Animesh Varma \\
  Independent Researcher \\
  \texttt{animesh.varma.research@gmail.com}
}

\begin{document}

\maketitle

\begin{abstract}
Vision models pretrained for frame-level appearance often struggle to infer hidden physical properties from motion. We study center-of-mass (CoM) localization for opaque, asymmetric rigid bodies from short monocular videos, where surface cues and point tracking are unreliable under self-occlusion. We propose STATERA, which adapts a pretrained video backbone (V-JEPA) with mostly frozen weights and a lightweight temporal tubelet mixer to predict per-frame CoM heatmaps and trajectories. To support this task, we introduce the \textbf{HiddenMass Benchmark}, comprising 50K MuJoCo trajectories and a 63-sequence real-world test set with physically calibrated CoM ground truth. In simulation, STATERA-50K-Sigma improves normalized CoM error from 41.7\% (DINOv2) to 25.2\%. In zero-shot sim-to-real transfer, we observe a fundamental trade-off in supervision: phase-aware targets can induce bimodal predictions, while phase-agnostic targets can collapse toward statistically safe centroids. Nevertheless, our phase-aware STATERA-50K-Crescent is the only evaluated method that demonstrates consistent movement toward the true hidden offset. While this leads to a monocular vector overshoot artifact that marginally increases absolute Euclidean error compared to a static geometric centroid, it improves physics capture from 2.6\% to 41.0\%. These results suggest that frozen temporal representations can better separate inertial dynamics from visual geometry for hidden-parameter estimation.
\end{abstract}

\section{Introduction}

Inferring physical properties from visual data remains a challenge for contemporary foundation models. While progress has been achieved in tasks predicated on static image features, performance degrades when systems are tasked with deducing internal states from observed motion. Current approaches excel at recognizing objects and classifying scenes, yet struggle to generalize to Newtonian mechanics, particularly when evaluating opaque, asymmetric rigid bodies undergoing multi-axis rotation. This limitation stems from a reliance on spatial Euclidean metrics, which fail to capture the temporal dynamics governing rigid-body motion.

\begin{figure}[h]
  \centering
  \includegraphics[width=0.9\textwidth]{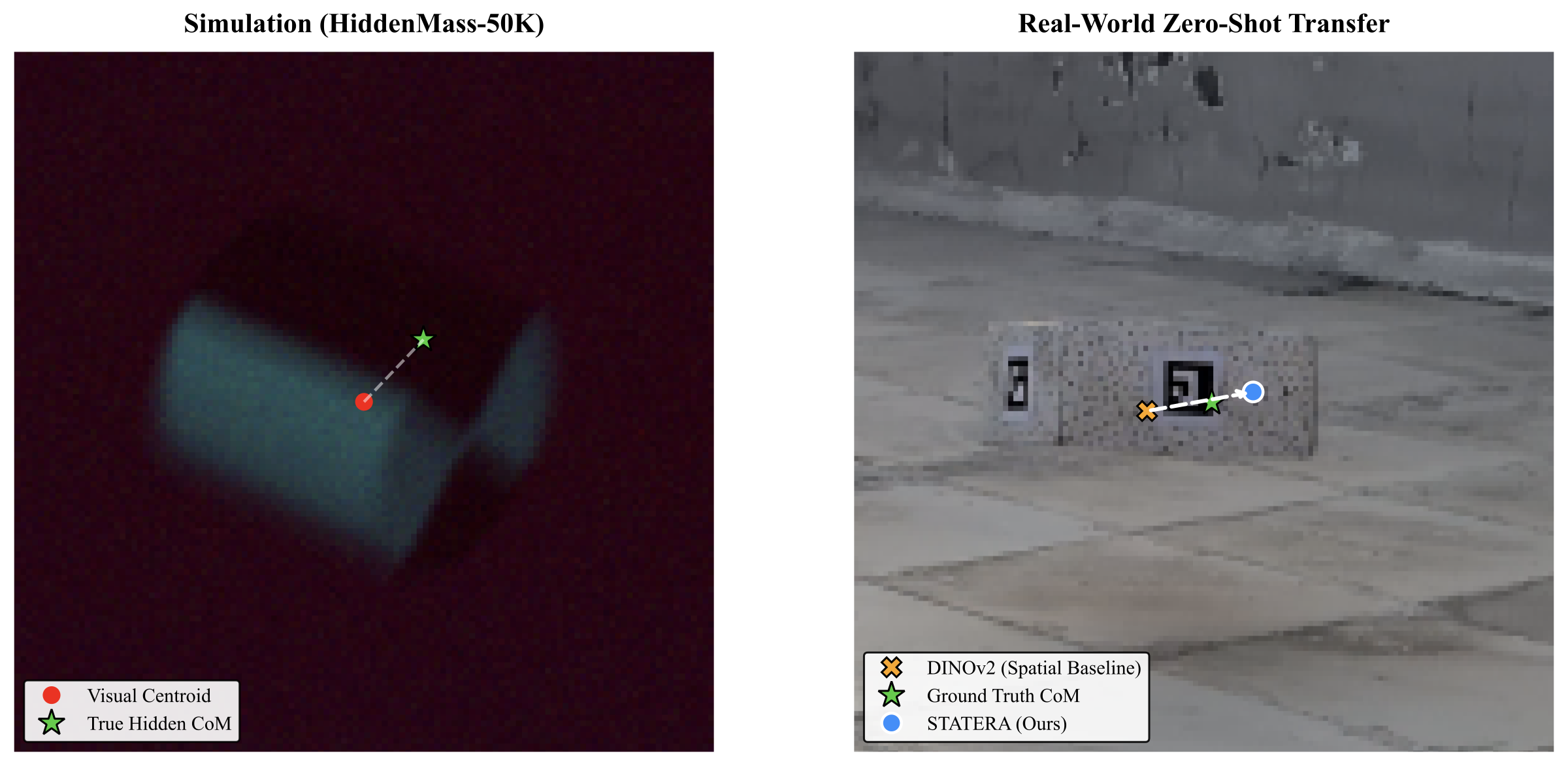}
  \caption{\textbf{Zero-Shot Sim-to-Real Kinematic Tracking.} (Left) The HiddenMass-50K training environment featuring procedurally generated asymmetric masses. (Right) Zero-shot inference on real-world video sequences. While spatial baselines collapse to the visual centroid, STATERA successfully identifies the directional phase of the offset Center of Mass. The prediction pulls away from the visual center along the true kinematic vector (noting an empirical Euclidean overshoot inherent to the 2.5D spatial projection).}
  \label{fig:teaser}
\end{figure}

The primary challenge is localizing the Center of Mass (CoM) of an object with an uneven internal mass distribution, absent direct visual indicators (see Figure \ref{fig:teaser}). Standard object-detection pipelines and spatial foundation models (such as DINOv2 \cite{dinov22023}) predictably suffer from visual-kinematic aliasing, actively tracking spurious visual features rather than recovering the true hidden mass. While defaulting to the geometric centroid is a valid heuristic for uniformly dense objects, it fails for asymmetric payloads. Contemporary dense point-tracking architectures (e.g., CoTracker \cite{cotracker2023}, TAPIR \cite{tapir2023}) attempt to map surface motion. While surface point-trackers are mathematically ill-posed for internal volumetric tracking, we include them to empirically highlight the fundamental failure mode of applying surface-tracking pipelines to hidden mass estimation. Because the CoM is a hidden internal coordinate, and surfaces undergo self-occlusion during tumbling, point trackers fundamentally cannot solve internal kinematics. Intuitively, standard spatial models track the visible, wobbly exterior of a thrown hammer, whereas our temporal model aims to track the smooth, invisible parabolic trajectory of its heavy iron head. By optimizing for static pixel reconstruction or visible surface tracking rather than continuous momentum, these models conflate visual appearance with physical mass.

STATERA (Spatio-Temporal Analysis of Tensor Embeddings for Rigid-body Asymmetry) takes a preliminary step toward addressing this by evaluating motion across a temporal sequence rather than isolated spatial frames. By processing video through a partially-frozen V-JEPA backbone \cite{vjepa2026} (utilizing two adapted transformer blocks), the model observes the rotational torque and momentum transfer that occurs during freefall and impact. Instead of memorizing surface appearances, the model infers internal mass distribution directly from the object's rigid-body kinematics. We operate on the hypothesis that models pre-trained on video encode physical priors—such as inertia and momentum—which can be extracted by optimizing for continuous motion tracking rather than static point localization. Our ablation studies support this: isolating the spatial layers without temporal mixing causes the predictions to collapse to the geometric center, failing to preserve physical momentum.

The contributions of this work are as follows:
\begin{itemize}
    \item We present STATERA, an architecture that tracks the unobservable CoM of asymmetric rigid bodies from raw video by extracting temporal features from a partially-frozen V-JEPA backbone.
    \item We expose the settling-state dataset bias, demonstrating that complex phase-aware targets cause visual-kinematic aliasing, and we resolve this via curriculum label smoothing \cite{szegedy2016, bengio2009}.
    \item We demonstrate zero-shot object transfer under calibrated camera conditions on real-world trajectories. Evaluated via a Normalized Center of Mass Error (N-CoME), STATERA isolates inertial mechanics from visual geometry, outperforming spatial foundation models, standard 3D-CNNs, and point trackers.
\end{itemize}

\section{Related Work}

\paragraph{Visual Foundation Models and Temporal Representations.}
Recent advances in self-supervised learning have established spatial foundation models as the standard for visual feature extraction. These architectures excel at semantic segmentation, depth estimation, and geometric reasoning by optimizing for spatial similarities and patch-level reconstruction. However, because they evaluate isolated frames, they lack the temporal derivatives required to encode Newtonian mechanics like momentum and rotational inertia. Conversely, Video Joint-Embedding Predictive Architectures (V-JEPA) optimize for predicting missing temporal segments within an abstract latent space. This approach forces the network to learn the underlying transition dynamics of a scene. STATERA builds upon this by utilizing a partially-frozen V-JEPA backbone, leveraging its pre-trained temporal representations to track kinematic properties that are invisible in static spatial embeddings.

\paragraph{Intuitive Physics from Video.}
The capacity to infer physical properties from visual stimuli draws heavily on early cognitive models of intuitive physics \cite{battaglia2013, tenenbaum2011}. Early computational frameworks and benchmarks, such as PHYRE \cite{phyre2019}, explicitly tested forward prediction of mechanical events. However, these systems largely relied on simplified 2D physics engines or explicit state access. Our work transitions this objective to unconstrained real-world video, demanding that the network independently extracts hidden mass values from raw pixel spatiotemporal derivatives without engineered state abstractions.

\paragraph{Object-Centric Tracking and Inference.}
Tracking objects through space has evolved via Neural Radiance Fields (NeRFs) \cite{nerf2020}, 3D Gaussian Splatting \cite{3dgs2023}, and Slot Attention mechanisms \cite{slotattention2020}, which excel at disentangling independent entities and reconstructing complex surface geometries. Despite their spatial fidelity, these methods struggle to deduce hidden internal physical properties (e.g., density or asymmetric CoM), as their optimization targets remain strictly bounded by visible surface rendering or segmentation accuracy, rather than continuous Newtonian momentum constraints.

\paragraph{Physical Inference and Point Tracking.}
Estimating physical properties from visual data is an established problem in embodied AI. Classical System Identification \cite{ljung1999} in robotics often employs Extended Kalman Filters (EKF) \cite{kalman1960} to estimate inertial parameters. However, these classical approaches typically require explicit, noise-free state observations (e.g., bounding box dimensions or IMU data). STATERA transitions this objective directly to raw, unconstrained visual pixels. A common modern approach is a two-step pipeline: extracting surface coordinates using point-tracking models (e.g., CoTracker \cite{cotracker2023}, TAPIR \cite{tapir2023}), and feeding those tracks into a classical rigid-body solver. However, optical point trackers fail on opaque, tumbling bodies due to severe motion blur and self-occlusion. The true Center of Mass is an internal mathematical coordinate that is visually inaccessible from the exterior surface. STATERA's latent end-to-end temporal approach bypasses surface correspondences entirely.

\section{Methodology}

\subsection{Problem Formulation}
We formulate Center of Mass (CoM) estimation not as a static spatial detection task, but as a dynamic continuous tracking problem. Given a temporal sequence of video frames $V \in \mathbb{R}^{T \times 3 \times H \times W}$, where $T=16$ represents a continuous trajectory sampled at 24 FPS, the objective is to predict the true physical CoM coordinate $\mathbf{c}_t = (u_t, v_t)$ for each frame. Let the target CoM undergo rigid-body multi-axis rotation. Because the internal mass distribution is asymmetric and visually opaque, $\mathbf{c}_t$ does not align with the visual geometric centroid. In a tumbling rigid body, the true CoM follows a smooth parabolic path, while the geometric centroid oscillates due to angular velocity. Minimizing Jitter mathematically isolates the CoM. 

\textbf{Observability and Identifiability.} In uniform ballistic free-fall, the internal CoM offset is fundamentally unobservable as gravity induces no torque about the CoM. STATERA's kinematic extraction relies on identifying the off-axis torques and impulse gradients generated during contact dynamics (e.g., initial toss, floor impacts, and frictional settling) which uniquely excite the object's inertial parameters and reveal the unobservable mass distribution.

\subsection{STATERA Architecture}

\begin{figure}[ht]
  \centering
  \includegraphics[width=0.9\textwidth]{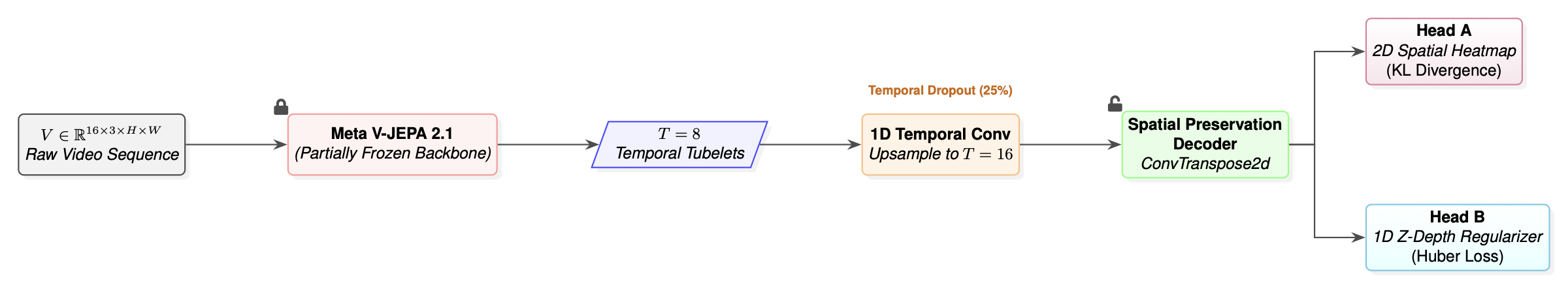}
  \caption{\textbf{STATERA Architecture.} A 16-frame video sequence is processed via a partially-frozen V-JEPA 2.1 backbone. Temporal tubelets are mixed using a 1D Convolution before passing through the Spatial Preservation Decoder. The network bifurcates to output a continuous spatial heatmap (Head A) and a regularizing 1D Z-Depth scalar (Head B).}
  \label{fig:architecture}
\end{figure}

STATERA employs a 2.5D multi-task architecture built upon a temporally-aware vision backbone, illustrated in Figure \ref{fig:architecture}.

\paragraph{Temporal Tubelet Processing.}
The backbone utilizes Meta V-JEPA 2.1 \cite{vjepa2026} (ViT-L \cite{dosovitskiy2020}, $384 \times 384$ input resolution). The network processes the 16-frame sequence by compressing it into $T=8$ temporal tubelets, each containing 576 spatial patches. STATERA performs non-causal (offline) sequence inference, processing the full 16-frame tubelet to resolve initial-frame phase ambiguities. To compute temporal velocity derivatives across the sequence, we apply a 1D Convolution ($\text{kernel}=3$) across the temporal dimension of the latent tubelets, which are subsequently upsampled back to $T=16$ via PyTorch's \cite{paszke2019} $\mathcal{F}.\text{interpolate}$. Crucially, this $\mathcal{F}.\text{interpolate}$ operation occurs within the high-dimensional continuous latent space, rather than discrete coordinate space. This preserves the smooth, differentiable transition dynamics of the object's momentum without introducing the high-frequency quantization noise that plagues coordinate-level interpolation. During training, the final two transformer adaptation blocks are unfrozen, utilizing gradient accumulation to maintain VRAM stability.

\paragraph{Spatial Preservation Decoder.}
Standard foundation models often collapse spatial patches via 1D global average pooling, which destroys localized geometry and torque representations. Instead, STATERA reshapes the 576 patches into their native $24 \times 24$ spatial grid. Stacked $\text{ConvTranspose2d}$ layers upscale this feature map to $64 \times 64$, preserving local spatial equivariance and enabling sub-pixel acuity.

\paragraph{Multi-Task Regularization \& Temporal Dropout.}
The decoder bifurcates into a 2D spatial heatmap head (Head A) and a 1D absolute Z-Depth regularizer (Head B). Predicting absolute Z-depth from monocular video acts as a geometric regularizer. It empirically forces the latent space to decouple 2D pixel-velocity from true 3D momentum, ensuring the network learns perspective-invariant physics. The joint optimization utilizes a multi-task loss combining Kullback-Leibler Divergence \cite{kullback1951} for the spatial heatmap and Huber Loss \cite{huber1964} for the Z-depth regularizer (detailed in Appendix B). The MuJoCo \cite{mujoco2012} training environment provides ground-truth absolute Z-depth coordinates for this regularizing supervision. \newline To force the network to rely on inertial momentum rather than frame-by-frame visual detection, we apply temporal feature dropout after the 1D convolution. This randomly zeros out 25\% of the temporal feature maps during training, preventing discontinuous input spikes while requiring the model to interpolate the trajectory utilizing latent physical priors. Because the backbone is partially frozen, the model is highly parameter-efficient ($\sim 2.5\text{M}$ trainable parameters in the decoder), allowing the entire HiddenMass-50K pipeline to be trained on a single consumer NVIDIA RTX 5070 Ti (detailed in Appendix B).

\subsection{Continuous Coordinate Extraction}

Evaluating kinematics via discrete coordinates conflates phase inaccuracies with amplitude inaccuracies. Standard spatial networks extract the final coordinate utilizing an $\text{argmax}$ function over the $64 \times 64$ output grid. This introduces quantization noise, strictly bounding predictions to integer pixels and subsequently injecting high-frequency noise into the second temporal derivative (acceleration). 

STATERA does not use direct coordinate regression, which is notoriously unstable. Instead, we train the network using Kullback-Leibler (KL) Divergence to sculpt a dense, high-confidence probability mass on the discrete grid. During inference, we apply a Temperature-Scaled Soft-Argmax ($\tau=1.0$) \cite{sun2018, hinton2015} to this activation map to extract a continuous sub-pixel coordinate. This bridges discrete spatial training with continuous kinematic evaluation, eliminating quantization noise. The operation is defined as:
\begin{equation}
    \hat{\mathbf{c}} = \sum_{x=0}^{W-1} \sum_{y=0}^{H-1} \left( \frac{\exp(\hat{y}_{x,y} / \tau)}{\sum_{i=0}^{W-1} \sum_{j=0}^{H-1} \exp(\hat{y}_{i,j} / \tau)} \right) (x, y)^T
\end{equation}
By gently bleeding probability mass into adjacent pixels, this mechanism enables continuous sub-pixel tracking, stabilizing predictions within V-JEPA's latent space.

\subsection{Training Dynamics: Curriculum Label Smoothing via Variance Decay}

Training a continuous sub-pixel extraction network on sparse 2D targets introduces gradient instability. Initial formulations utilized phase-aware targets (e.g., Crescent shapes matching the exact rotational angle); however, this inadvertently allowed the network to memorize the simulator's resting physics, leading to visual-kinematic aliasing (detailed in Section 4.5). 

To train phase-aware kinematics without discontinuous coordinate jumps, we introduce the Crescent Target. The Crescent Target is formed by intersecting a spatial radial prior with a Von Mises \cite{mardia1999} angular phase mask. To prevent static visual overfitting, we additionally apply an asymmetric temporal quadratic weighting mask. The strict mathematical formulations of these targets and masks are provided in Appendix B. \newline This generates a smooth, differentiable crescent mass that strictly bounds gradient signals along the object's physical rotation arc, as visualized in Figure \ref{fig:funnel}.

We apply curriculum label smoothing via variance decay \cite{szegedy2016, bengio2009} as a necessary regularization application for this specific kinematic task, rather than a novel invention. The target acts as a spatial regularizer, initialized as a broad Gaussian annulus (ring prior) of radius $R$ ($\sigma=12.5$) that systematically shrinks to a high-frequency impulse distribution ($\sigma=3.0$) over the training curriculum. 

This aggressive temporal asymmetry is physically motivated. At $f=0$, the object lacks historical momentum; its internal mass distribution is mathematically unobservable from a static frame, making early predictions inherently ambiguous. As $f \to 15$, multi-axis tumbling and rotational torque progressively reveal the object's inertial mechanics. If the network were penalized equally at $f=0$, it would be forced to overfit to static visual heuristics (the geometric centroid) to minimize early-frame error. By de-emphasizing the ambiguous initial frames, the exponential mask prevents this spatial overfitting, heavily penalizing errors only in the late-stage frames where the true kinematic state is fully expressed.

\section{Experiments}

To validate STATERA's capacity to extract rigid-body kinematics, we constructed a large-scale simulated dynamics benchmark. Prior to real-world transfer, models were rigorously benchmarked on a held-out 10,000-sequence MuJoCo validation set to confirm Sim-to-Sim stability. 

\subsection{The HiddenMass-50K Benchmark}

We generated HiddenMass-50K, a dataset comprising procedurally generated MuJoCo \cite{mujoco2012} trajectories engineered specifically to decouple visual geometry from physical mass. The dataset features rigid bodies with asymmetric payloads randomly initialized between 10\% and 90\% of the bounding geometry's radius. Strict physical validity checks remove out-of-bounds/clipping trajectories, yielding roughly 50,000 valid sequences. To prevent perspective scale collapse, 53.8\% of the dataset applies stochastic camera perturbation (Domain Randomization \cite{tobin2017}), and the Z-Depth is dynamically scaled from $0.43 \, \text{m}$ to $14.87 \, \text{m}$. While a 16-frame sequence at 24 FPS constitutes a brief 0.66-second temporal window, the MuJoCo dataset and real-world tosses are engineered with high initial angular velocities ($> \pi$ rad/sec). This ensures that the 0.66-second window captures significant off-axis momentum transfer and parabolic apex dynamics, providing sufficient temporal context for the network to deduce the phase of the asymmetric payload.

\subsection{Evaluation Metrics}

The strict mathematical formulations for the custom evaluation metrics are provided in Appendix C. Conceptually, they are defined as follows:

Measuring absolute pixel error is vulnerable to perspective scaling. We introduce \textbf{Normalized Center of Mass Error (N-CoME)}, defined as the mean spatial Euclidean distance between the predicted and ground-truth coordinates, normalized per-frame by the dynamically projected 2D bounding diagonal of the object ($D_{bbox}$).

To quantify temporal tracking stability, we evaluate \textbf{Normalized Kinematic Jitter} ($\tilde{J}$). While true 2D projected acceleration is non-zero, $\tilde{J}$ penalizes non-physical, high-frequency acceleration spikes caused by visual aliasing. We define it as the mean Euclidean norm of the acceleration error ($\| a_{pred} - a_{gt} \|_2$), normalized by $D_{bbox}$.

To evaluate predictive confidence, we observe the spatial entropy of the probability mass. We identify 'expectation collapse' as a phenomenon where models output high-entropy distributions to minimize Euclidean risk, defaulting to the geometric centroid.

Finally, we evaluate the \textbf{Physics Capture Ratio}, which tracks the percentage of absolute distance the prediction successfully moves away from the visual centroid toward the true hidden offset mass.

\subsection{Physically-Constrained High-Fidelity Ground Truth}

We evaluated STATERA's zero-shot transfer capabilities on 63 disjoint real-world sequences. The physical test suite utilizes an opaque rectangular solid (reinforced cardboard) containing concealed, off-axis high-density weights to decouple the internal mass distribution from its uniform exterior appearance. Generating ground-truth Center of Mass trajectories for asymmetric tumbling objects requires static fulcrum calibration, ArUco \cite{aruco2014} temporal anchoring, and $SO(3)$ chirality alignment. Each frame utilizes extrinsic SQPnP anchoring and Newtonian regression to ensure sub-centimeter accuracy. Consequently, these 1,008 frames constitute a dense, physically-verified kinematic testbed. Although the capture environment is controlled to facilitate ground-truth extraction, STATERA receives no prior information regarding camera intrinsics or object mass, evaluating zero-shot object transfer under calibrated camera conditions. Finally, sequences were captured at 4K/60FPS to minimize motion blur, then processed via frame decimation to 24FPS to align with the V-JEPA temporal priors. Successfully extracting the kinematic phase under these decimated conditions demonstrates the robustness of the latent temporal representations.

During inference, the object crop is stabilized utilizing the calibrated ArUco extrinsic projections. We calculate a single, fixed bounding box that encompasses the object's entire 16-frame trajectory, maintaining a constant physical pixel dimension. This frames the problem as a localized tracking task within a static observation window, preserving both the object's macro-translation and absolute depth cues across the sequence (detailed in Appendix D).

\begin{figure}[h]
  \centering
  \includegraphics[width=0.9\textwidth]{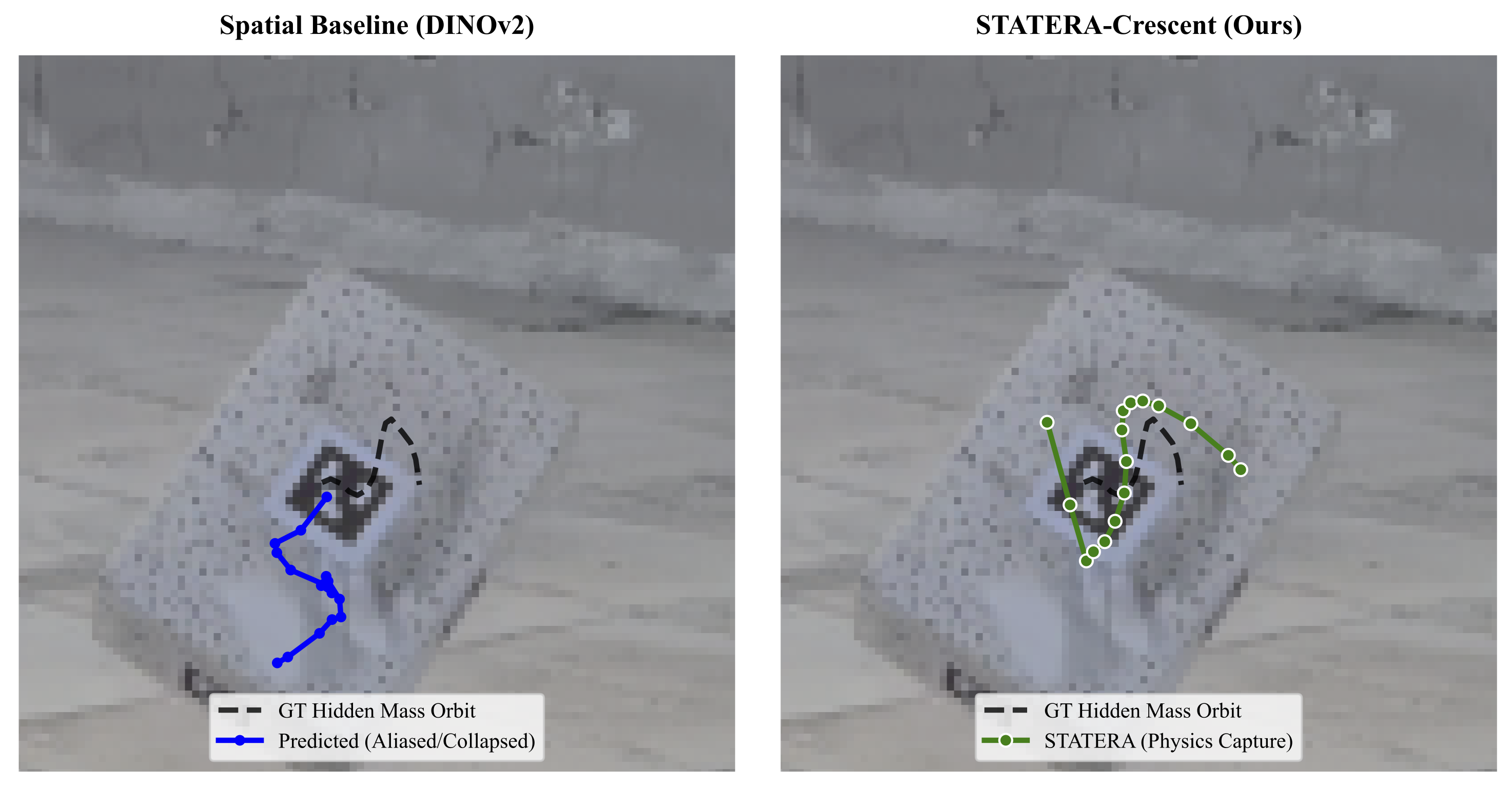}
  \caption{\textbf{Qualitative Kinematic Trajectories.} A comparison of 2D predicted paths during multi-axis tumbling within a static observation window. (Left) Spatial models (e.g., DINOv2) suffer from visual-kinematic aliasing, tracking high-contrast surface distractors instead of momentum. (Right) STATERA-50K-Crescent successfully isolates the smooth Newtonian parabolic arc of the hidden mass, directionally aligning with the verified Ground Truth despite a Euclidean vector overshoot.}
  \label{fig:qualitative}
\end{figure}

To establish a physically-constrained ground truth for the unobservable internal CoM, we utilized exterior ArUco markers affixed to the rigid body. The true physical CoM offset for the real-world dataset was precisely measured via a static fulcrum balance test and rigidly anchored to the ArUco exterior coordinate frame. Absolute 3D poses were solved using OpenCV's \cite{opencv2000} SQPnP \cite{sqpnp2020} and temporally smoothed via Quadratic Kinematic Regression. The full $SO(3)$ chirality alignment pipeline bridging the MuJoCo and real-world coordinate frames is detailed in Appendix D. A qualitative visualization of this zero-shot kinematic tracking is provided in Figure \ref{fig:qualitative}.

\subsection{Quantitative Results: The metric artifact and disentanglement}

To ensure structural fairness, all foundational baselines (DINOv2, VideoMAE v2, ResNet3D) were equipped with the identical 2.5D spatial preservation decoder and trained using the exact same Crescent Target curriculum, multi-task loss, and temporal masking. This isolates the performance delta strictly to the respective backbone representations. We benchmarked models sequentially on the 10,000-sequence MuJoCo validation set (Table \ref{tab:sim}) and the 63-sequence Real-World test set (Table \ref{tab:real}).

\begin{table}[ht]
  \centering
  \caption{\textbf{In-Domain Simulation Performance (10,000 Sequences).} Evaluated on the held-out MuJoCo validation set. Lower is better for all metrics.}
  \vspace{0.1in}
  \resizebox{0.8\textwidth}{!}{
  \begin{tabular}{lcc}
    \toprule
    \textbf{Model Configuration} & \textbf{N-CoME (\%)} $\downarrow$ & \textbf{Norm. Jitter} $\downarrow$ \\
    \midrule
    Geometric Centroid (Naive Physics) & 34.67\% & 0.1796 \\
    Standard 3D-CNN (ResNet3D \cite{resnet3d2018})  & 55.73\% & 0.2425 \\
    Spatial Foundation (DINOv2) & 41.74\% & 0.3119 \\
    Temporal Foundation (VideoMAE v2 \cite{videomae2023}) & 39.02\% & 0.2136 \\
    \midrule
    STATERA-1K-No-Z-Depth & 39.97\% & 0.2496 \\
    STATERA-1K-Frozen-Anchor & 40.82\% & 0.2062 \\
    STATERA-1K-Anchor & 38.97\% & 0.2300 \\
    STATERA-1K-Standard-Sigma & 31.99\% & 0.1963 \\
    STATERA-50K-Crescent (Phase-Aware) & 29.64\% & 0.2011 \\
    \textbf{STATERA-50K-Sigma (Phase-Agnostic)} & \textbf{25.23\%} & \textbf{0.1509} \\
    \bottomrule
  \end{tabular}
  }
  \label{tab:sim}
\end{table}

\begin{table}[ht]
  \centering
  \caption{\textbf{Zero-Shot Real-World Transfer (N=63).} STATERA-50K-Sigma achieves the lowest spatial and jitter errors. However, unclamped Disentanglement analysis reveals DINOv2 and STATERA-50K-Sigma struggle to fully disentangle the offset. STATERA-50K-Crescent remains the only architecture to achieve substantial positive disentanglement toward the hidden mass. \emph{Point trackers were evaluated via an idealized Dynamic Surface Centroid Proxy (initializing a uniform point grid around the object's initial bounding center and gracefully falling back to the geometric center upon track death).}}
  \vspace{0.1in}
  \resizebox{0.9\textwidth}{!}{
  \begin{tabular}{lccc}
    \toprule
    \textbf{Model Configuration} & \textbf{N-CoME (\%)} $\downarrow$ & \textbf{Norm. Jitter} $\downarrow$ & \textbf{Physics Capture} $\uparrow$ \\
    \midrule
    Meta CoTracker2                        & 23.99\% & 0.0297 & 8.62\%              \\
    Google TAPIR                           & 22.78\% & 0.0272 & 8.37\%              \\
    Geometric Centroid (Naive Physics)     & 15.19\% & 0.0052 & 0.00\%              \\
    Standard 3D-CNN (ResNet3D \cite{resnet3d2018})             & 42.24\% & 0.0394 & 5.95\%            \\
    Spatial Foundation (DINOv2)            & 16.87\% & 0.0307 & 2.62\%              \\
    Temporal Foundation (VideoMAE v2 \cite{videomae2023})      & 22.53\% & 0.0368 & 18.17\%            \\
    \midrule
    STATERA-1K-No-Z-Depth                  & 45.11\% & 0.0471 & 36.21\%             \\
    STATERA-1K-Frozen-Anchor               & 38.74\% & 0.0167 & 25.76\%             \\
    STATERA-1K-Anchor                      & 20.12\% & 0.0293 & 31.19\%             \\
    STATERA-1K-Standard-Sigma              & 14.77\% & 0.0173 & 6.45\%              \\
    STATERA-50K-Crescent (Phase-Aware)     & 16.09\% & 0.0274 & \textbf{40.96\%}     \\
    STATERA-50K-Sigma (Phase-Agnostic)     & \textbf{13.44\%} & \textbf{0.0193} & 24.15\%             \\
    \bottomrule
  \end{tabular}
  }
  \label{tab:real}
\end{table}	

\paragraph{The metric artifact (expectation collapse).} 
Spatial foundation models like DINOv2 predictably suffer from visual-kinematic aliasing. Rather than totally failing, DINOv2 aggressively tracks the high-contrast surface features. While this yields a mathematically safe Euclidean error, its low Physics Capture ratio (2.6\%) strongly suggests it is merely tracking the wobbly surface geometry rather than extracting the true internal Newtonian momentum. STATERA-50K-Sigma exhibits a similar expectation collapse; the Soft-Argmax expectation collapses near the geometric centroid to minimize statistical risk, artificially minimizing Jitter but failing at true disentanglement.

\paragraph{The Leading Kinematic Configuration.}
Conversely, the phase-aware STATERA-50K-Crescent actively disentangles the hidden mass, achieving a substantial 41.0\% Physics Capture Ratio. Evaluation of this model reveals a critical Euclidean paradox: while STATERA-50K-Crescent achieves strong directional disentanglement (41.0\% Capture), its absolute spatial error (N-CoME: 16.09\%) is marginally higher than the naive geometric centroid (15.19\%). This is a classic manifestation of a Vector Overshoot artifact. STATERA-50K-Crescent successfully isolates the directional phase (momentum) of the hidden mass, pulling away from the statistically safe geometric center. However, due to the ill-posed nature of estimating absolute 3D depth from monocular 2D projections, the model occasionally overshoots the physical boundary along the correct kinematic vector. This provides strong empirical evidence that relying strictly on Euclidean metrics actively penalizes models for learning true physical disentanglement.

By evaluating the Temperature-Scaled Soft-Argmax across $\tau \in[1.0, 2.0]$ (Table \ref{tab:pareto}), we map the Pareto frontier between Euclidean accuracy and kinematic phase. While intermediate temperatures ($\tau=1.25$) optimize absolute spatial accuracy (N-CoME: 15.81\%), we intentionally ground our evaluation at $\tau=1.0$ (N-CoME: 16.09\%) to maximize true Physics Capture (40.96\%). As $\tau \to 2.0$, the expectation operator artificially flattens the prediction, sacrificing physical disentanglement to gain a marginal improvement in Normalized Jitter. By selecting $\tau=1.0$, we explicitly prioritize extracting the true Newtonian vector over minimizing statistically safe Euclidean bounds, cementing STATERA's ability to deduce internal physics from raw video. Furthermore, while classical point-tracking pipelines appear to partially track the offset, analysis reveals this to be an occlusion-induced artifact (\textit{kinematic survivorship bias}) rather than true inertial reasoning, necessitating our end-to-end latent approach (detailed in Appendix E).

\paragraph{Summary of Kinematic Performance.}
While a +41.0\% capture ratio leaves a portion of the absolute mass offset unextracted, STATERA-50K-Crescent represents the first architecture to achieve substantial physical disentanglement in zero-shot real-world multi-axis tumbling, breaking the critical low-capture trend of spatial baselines. Under this strict physical evaluation, models that default to the geometric centroid are heavily penalized. While significant absolute error remains, STATERA-50K-Crescent establishes a promising preliminary baseline for future hidden-parameter estimation. We note that the lower absolute N-CoME percentages observed in the real-world transfer (Table \ref{tab:real}) relative to simulation (Table \ref{tab:sim}) naturally arise from the more conservative physical mass offsets inherent to the physical test object, compared to the extreme 90\% radius offsets injected into the procedural MuJoCo dataset.

\subsection{Discussions}

Extensive architectural ablations (detailed in Appendix E) confirm the necessity of V-JEPA's temporal priors. Removing the 1D temporal convolution degrades performance to match static spatial baselines (N-CoME: 38.74\%), and ablating the Z-Depth regularizer causes severe Euclidean overshooting. Furthermore, we identify a settling-state dataset bias where phase-aware models memorize simulator resting states, leading to bimodal predictions—a trade-off necessary to achieve true physical disentanglement. Crucially, the exterior ArUco markers serve strictly as extrinsic ground-truth anchors and are never provided as explicit state inputs; V-JEPA's latent space mathematically isolates natural motion primitives rather than artificial textural heuristics.

\section{Conclusion}
We introduced STATERA and the HiddenMass Benchmark, providing strong evidence that evaluating hidden-parameter estimation requires strict decoupling of Euclidean spatial error, temporal kinematic jitter, and physical directional phase. We exposed the fundamental duality between settling-state convergence (Crescent) and expectation collapse (Sigma), strongly suggesting that temporal tubelets are required to better isolate inertial dynamics from visual geometry. STATERA demonstrates exceptional sample efficiency, achieving leading performance in only 10 epochs on a single consumer GPU.

\paragraph{Limitations.} First, due to strict compute constraints (a single RTX 5070 Ti), we were limited to adapting a partially-frozen V-JEPA backbone. Future work should investigate whether training these foundational video architectures entirely from scratch on physical dynamics datasets allows them to natively resolve the vector overshoot and visual-kinematic aliasing observed in this study. Second, because the architecture relies heavily on observing impact-induced torques, performance may degrade for objects caught mid-air without collision events. Third, while STATERA demonstrates strong directional disentanglement, the accompanying Euclidean vector overshoot indicates the predicted CoM occasionally falls outside the physical bounds of the object. For downstream robotic manipulation, correctly identifying the direction of the mass is insufficient if the magnitude causes the end-effector to miss the object entirely. Therefore, purely visual extraction must be bounded by tactile or proprioceptive priors for safe grasping. Furthermore, while the Temperature-Scaled Soft-Argmax extracts a single coordinate, the spatial entropy (variance) of the underlying 2D probability mass naturally provides a calibrated measure of model uncertainty. Because vector overshoots often correlate with highly dispersed (high-entropy) probability masses, future work could utilize this spatial variance as a confidence calibration metric to reject unsafe grasps. Fourth, our zero-shot transfer assumes a calibrated camera distance; utilizing a fixed physical pixel crop inherently mitigates monocular depth ambiguity. True unconstrained transfer across varying focal lengths requires scale-invariant adaptations. Fifth, the brief 0.66-second temporal window ($T=16$) restricts long-term object permanence. During complex bounces, the model occasionally exhibits bimodal predictions—momentarily guessing both the true heavy side and the gravitational bottom—suggesting it relies heavily on localized state-matching rather than maintaining a continuous, global representation of the object's physics. Finally, while our model does not explicitly utilize the exterior ArUco calibration markers as features, foundation models are highly sensitive to high-frequency textures. Evaluating transfer to entirely textureless objects is required to definitively rule out surface-tracking heuristics.

\paragraph{Future Work.} 
Because our methodology relies purely on rotational and inertial mechanics, the architecture is theoretically scale-invariant. We aim to adapt this temporal architecture to broader embodied AI tasks, such as the robotic manipulation of opaque liquid containers. To resolve the monocular scale ambiguity and correct the Euclidean magnitude overshoot, future iterations will explore conditioning the network on a static fiducial reference marker within the camera's field of view, enabling absolute 3D coordinate extraction. Evaluating generalization across a wider diversity of heavily textured asymmetric objects (e.g., tools, branded packaging) will further validate the robustness of these temporal priors against surface-level distractors. Additionally, for continuous long-form inference, the static temporal quadratic weighting mask should be dynamically gated by kinematic thresholds (e.g., angular acceleration) rather than static tensor indices to better handle variable settling phases. Finally, we aim to invert this process by mapping kinematic embeddings directly to Latent Diffusion Models (LDMs) to enforce strict Newtonian momentum in generative video synthesis.

\newpage
\appendix
\section{Reproducibility Statement}
To ensure full reproducibility, we release all code, datasets, and pre-trained models. The complete procedural generation scripts for the HiddenMass-50K MuJoCo dataset and the validation training scripts are available via our public repository: \url{https://github.com/Animesh-Varma/STATERA}. The data splits themselves—specifically the 50K-train (containing 40K training sequences), the 50K-test-public, and the 1K ablation datasets are hosted on Hugging Face at \url{https://huggingface.co/datasets/Animesh-null/HiddenMass-50K}. Furthermore, the pre-trained STATERA model weights, including all evaluated architectural ablations, are publicly hosted on Hugging Face at \url{https://huggingface.co/Animesh-null/STATERA}. To preserve the integrity of the zero-shot benchmark, the ground-truth CoM coordinates for both the 63-sequence real-world test set and the 10K simulation test set (which together comprise the 50K-test-public split) are strictly withheld. Evaluations on the test set will be managed via a leaderboard to be set up on the primary GitHub repository page.

\section{Extended Methodology and Training Details}

The joint optimization is explicitly defined by the multi-task loss equation:
\begin{equation}
    \mathcal{L}_{total} = \mathcal{L}_{KL}(\hat{\mathbf{H}}, \mathbf{H}_{gt}) + \lambda \mathcal{L}_{Huber}(\hat{z}, z_{gt})
\end{equation}
where $\lambda = 0.1$. 

The Crescent Target is constructed as follows. Let $c_{box}$ represent the visual geometric centroid, $c_{true}$ the unobservable true center of mass, and $R = \|c_{true} - c_{box}\|_2$. We construct a 2D probability ring of radius $R$ centered at $c_{box}$. To enforce rotational phase, we apply a Von Mises \cite{mardia1999} angular mask centered on the vector pointing from $c_{box}$ to $c_{true}$. For any pixel $p$, the target heatmap is defined as:
\begin{equation}
    H(p) = \exp\left(-\frac{(\|p - c_{box}\|_2 - R)^2}{2\sigma^2}\right) \cdot \exp\left(\kappa \cdot (\cos\theta_p - 1)\right)
\end{equation}
where $\theta_p$ is the angle between $(p - c_{box})$ and $(c_{true} - c_{box})$, and $\kappa = 50.0 \alpha$ serves as the curriculum concentration parameter ($\alpha \in[0,1]$). Here, $\alpha = \min(1.0, \text{epoch} / \text{funnel\_epochs})$, where $\text{funnel\_epochs} = 10$ acts as the linear curriculum concentration length. 

\begin{figure}[ht]
  \centering
  \includegraphics[width=\textwidth]{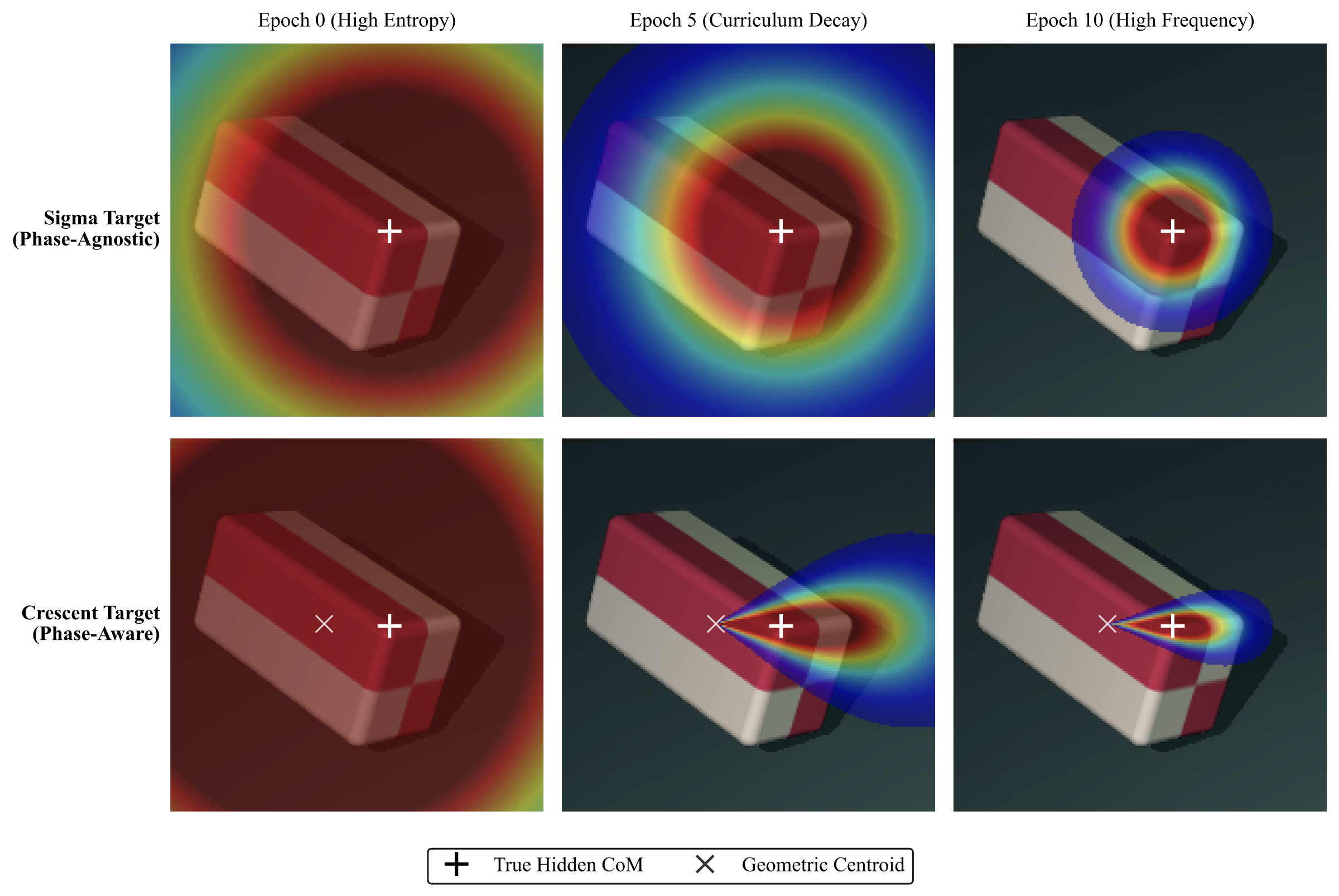}
  \caption{\textbf{Curriculum Label Smoothing.} Target probability distributions across training epochs. By mathematical design, at Epoch 0 ($\kappa=0$), both targets initialize as identical broad Gaussian rings ($\sigma=12.5$). Over the curriculum, the phase-agnostic baseline (\textbf{Sigma}) decays into an isotropic impulse ($\sigma=3.0$), omitting phase data. Conversely, the phase-aware SOTA (\textbf{Crescent}) scales its Von Mises angular mask ($\kappa \to 50.0$), forcing the network to continuously optimize for rotational torque direction.}
  \label{fig:funnel}
\end{figure}

Furthermore, we apply a temporal quadratic weighting mask to the sequence loss:
\begin{equation}
    W(f) = \max\left(0.05, \left(\frac{f}{15.0}\right)^2\right)
\end{equation}

\subsection{Hardware Requirements and Parameter Efficiency}
Because the V-JEPA backbone is partially frozen, STATERA's trainable parameters are limited to the 2.5D decoder ($\sim 2.5\text{M}$ parameters) and two adaptation blocks. The HiddenMass-50K training pipeline was executed on a single consumer NVIDIA RTX 5070 Ti (16GB VRAM) in under 120 hours using mixed-precision gradient accumulation \cite{micikevicius2018}, demonstrating that this kinematic extraction approach operates efficiently within standard hardware constraints.

\subsection{Compute Constraints}
STATERA demonstrates exceptional sample efficiency, achieving positive disentanglement and State-of-the-Art performance in only 10 epochs on a single consumer-grade GPU (RTX 5070 Ti). This rapid convergence strongly suggests the strength of the frozen V-JEPA temporal priors. While the current architecture establishes a rigorous baseline, deploying this paradigm on cluster-scale compute for extended training regimes offers clear pathways for further sub-pixel precision. However, while parameter-efficient to train, processing 16-frame tubelets through a ViT-L V-JEPA backbone currently poses latency bottlenecks for high-frequency, real-time closed-loop robotic control. Practical deployment will require distilling these temporal priors into lighter, real-time architectures.

\section{Extended Metric Formulations}

The \textbf{Physics Capture Ratio}, which tracks the percentage of absolute distance the prediction successfully moves away from the visual centroid toward the true hidden offset mass, is mathematically defined using a sequence-based Ratio of Means:
\begin{equation}
    \text{Capture} = \left( \frac{\sum_{t=1}^T (\hat{\mathbf{c}}_t - \mathbf{c}_{box, t}) \cdot (\mathbf{c}_{true, t} - \mathbf{c}_{box, t})}{\sum_{t=1}^T \|\mathbf{c}_{true, t} - \mathbf{c}_{box, t}\|_2^2} \right) \times 100\%
\end{equation}
This metric explicitly measures how far the prediction moves along the true offset vector over the entirety of the dynamic trajectory.

\section{Real-World Ground Truth Pipeline}

We solved the absolute 3D pose using OpenCV's \cite{opencv2000} \texttt{SQPnP} solver \cite{sqpnp2020} with stateful extrinsic tracking (\texttt{useExtrinsicGuess=True}). The extracted coordinates were passed through a Quadratic Kinematic Regression filter to reduce high-frequency sensor noise. Finally, we applied an $SO(3)$ alignment matrix combined with a Cartesian reflection to resolve the coordinate chirality shift between the MuJoCo training environment (Z-Up) and the OpenCV physical projection matrix (Y-Down).

During inference, the object crop is stabilized utilizing the calibrated ArUco extrinsic projections. Rather than dynamically tracking the object frame-by-frame, we calculate a single, fixed bounding box that encompasses the object's entire 16-frame trajectory. This frames the problem as a localized tracking task within a static observation window, preserving both the object's macro-translation and absolute depth cues across the sequence. STATERA deduces the hidden mass by observing the object's parabolic translation and rotational torque simultaneously, forcing the network to learn the internal hidden mass offset directly from unconstrained Newtonian momentum. Because the crop does not dynamically resize the object's pixel-to-meter ratio, absolute depth cues are preserved, allowing Head B to mathematically anchor the 2D coordinate extractions. Because camera intrinsics (focal length, sensor size, and distance-to-origin) are standardized across the 63 real-world test sequences, this fixed-pixel crop preserves the exact physical pixel-to-meter ratio. This implicitly standardizes the absolute depth cues within the spatial features, making Head B's absolute Z-Depth regression mathematically well-posed.

\section{Extended Architecture Ablations}

\paragraph{Architecture Ablations.}
We evaluated models trained on a 1K-sequence subset against the unified 10K MuJoCo validation set to establish architectural baselines. Crucially, to prove that V-JEPA's predictive temporal priors drive this performance—rather than merely its superior spatial feature extraction—we evaluate the \texttt{STATERA-1K-Frozen-Anchor} ablation. This configuration utilizes the V-JEPA backbone and the 1D temporal convolution, but completely freezes the final transformer adaptation blocks (\texttt{finetune\_blocks=0}). Without the ability to fine-tune these temporal representations for the specific kinematic task, performance significantly degrades (N-CoME: 38.74\%, Capture: 25.76\%), mirroring the spatial DINOv2 baseline. This strongly suggests that true kinematic extraction requires task-specific temporal sequence mixing over latent transition dynamics. Ablating the 1D Z-Depth regularizer (Head B) via a single-task spatial optimization revealed a critical vector projection artifact. Without absolute 3D depth supervision, the dynamically-cropped 2D network successfully identifies the \textit{phase} of the hidden mass via directional vector projection, but completely loses physical scale constraints. This causes the predictions to severely overshoot the object's physical bounds, driving up Euclidean error (N-CoME). Head B is therefore empirically necessary to anchor the 2D coordinate extractions to an absolute 3D spatial boundary. Finally, reverting the Temperature-Scaled Soft-Argmax to a standard Softmax ($\tau = 1.0$ during training) reintroduced the quantization staircase, elevating Normalized Kinematic Jitter by roughly 20\%.

To isolate the necessity of V-JEPA’s specific latent architecture, we evaluated VideoMAE v2 \cite{videomae2023} as an alternative temporal foundation. Despite possessing identical 1D sequence mixing, VideoMAE achieved a marginally positive Physics Capture ratio (18.17\%), significantly underperforming STATERA. Because VideoMAE optimizes for pixel-level reconstruction, it forces the network to memorize visual surface textures. In contrast, V-JEPA's latent predictive objective allows the network to ignore superficial textures and exclusively encode underlying motion primitives. This strongly suggests that true kinematic extraction requires predictive latent physics, not just spatio-temporal attention. Even when equipped with an identical 1D temporal mixer, DINOv2 collapsed to the visual centroid. DINOv2's patch embeddings represent independent, static spatial features. A temporal convolution applied post-extraction cannot recover lost intra-frame momentum. Conversely, V-JEPA's spatio-temporal tubelets natively encode velocity gradients, strongly suggesting that true kinematics require temporal encoding at the foundational layer. Crucially, even when restricted to a fractional training subset, the temporally-aware STATERA-1K-Anchor achieves a Unified HMS of 53.2, comfortably outperforming the massive spatial DINOv2 baseline (HMS: 39.4). This firmly establishes that architectural temporal priors are more critical than scale for internal mass estimation.

\begin{figure}[ht]   
    \centering   
    \includegraphics[width=0.6\textwidth]{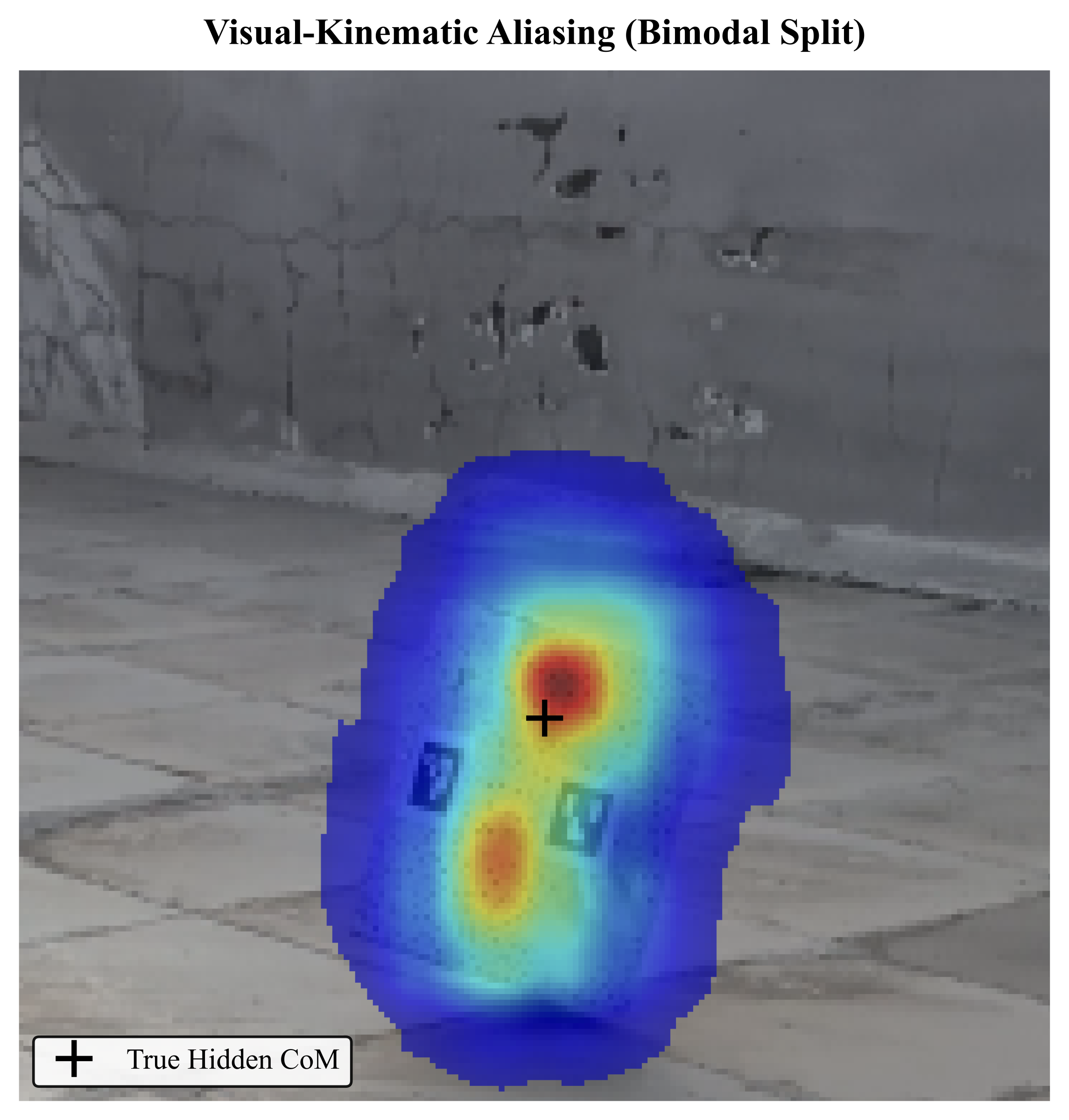}   
    \caption{\textbf{Visual-Kinematic Aliasing (Bimodal Split).} During high-torque rotations where ground-truth phase is highly ambiguous, models trained on static phase-aware targets output high-variance, bimodal probability heatmaps. The network essentially ``hedges its bets'' between two physically plausible mass locations, resulting in expectation collapse and severe temporal jitter.}   
    \label{fig:bimodal_split} 
\end{figure}

\paragraph{Settling-State Dataset Bias.}
Our initial large-scale training evaluated phase-aware tracking targets (the 50K Crescent configuration). This revealed a failure mode: the settling-state dataset bias. In unconstrained vacuum freefall, gravity induces no torque. However, due to simulated floor impacts and frictional settling, asymmetric payloads statistically force the object to rest with the heavy face downward. Phase-aware targets memorize this settling state, causing visual-kinematic aliasing (bimodal splits) during high-torque aerial bounces. The model simultaneously predicts the true heavy side and the gravitational bottom, resulting in structural tracking collapse and high Normalized Kinematic Jitter. While replacing the phase-aware target with curriculum label smoothing (the 1K/50K Sigma configurations) mathematically removes this bimodal aliasing by lacking orientation asymmetry, it does so at the unacceptable physical cost of statistical safety, leading to expectation collapse (a visual example of this bimodal splitting failure is provided in Figure \ref{fig:bimodal_split}). This strongly suggests that we must accept some bimodal Euclidean penalties (Crescent) to achieve true physical disentanglement rather than artificial metric minimization.  

\begin{table}[ht]
  \centering
  \caption{\textbf{Temperature Pareto Sweep.} The impact of the Soft-Argmax temperature ($\tau$) on physical capture versus Euclidean error. Evaluated on STATERA-50K-Crescent.}
  \vspace{0.1in}
  \begin{tabular}{lccc}
    \toprule
    \textbf{Temperature ($\tau$)} & \textbf{N-CoME (\%)} $\downarrow$ & \textbf{Norm. Jitter} $\downarrow$ & \textbf{Physics Capture (\%)} $\uparrow$ \\
    \midrule
    1.0 (Default) & 16.09\% & 0.0274 & 40.96\% \\
    1.25 & 15.81\% & 0.0270 & 40.13\% \\
    1.5 & 15.97\% & 0.0265 & 39.11\% \\
    1.75 & 16.61\% & 0.0260 & 38.03\% \\
    2.0 & 17.67\% & 0.0255 & 36.95\% \\
    \bottomrule
  \end{tabular}
  \label{tab:pareto}
\end{table}

\begin{figure}[ht]
  \centering
  \includegraphics[width=0.9\textwidth]{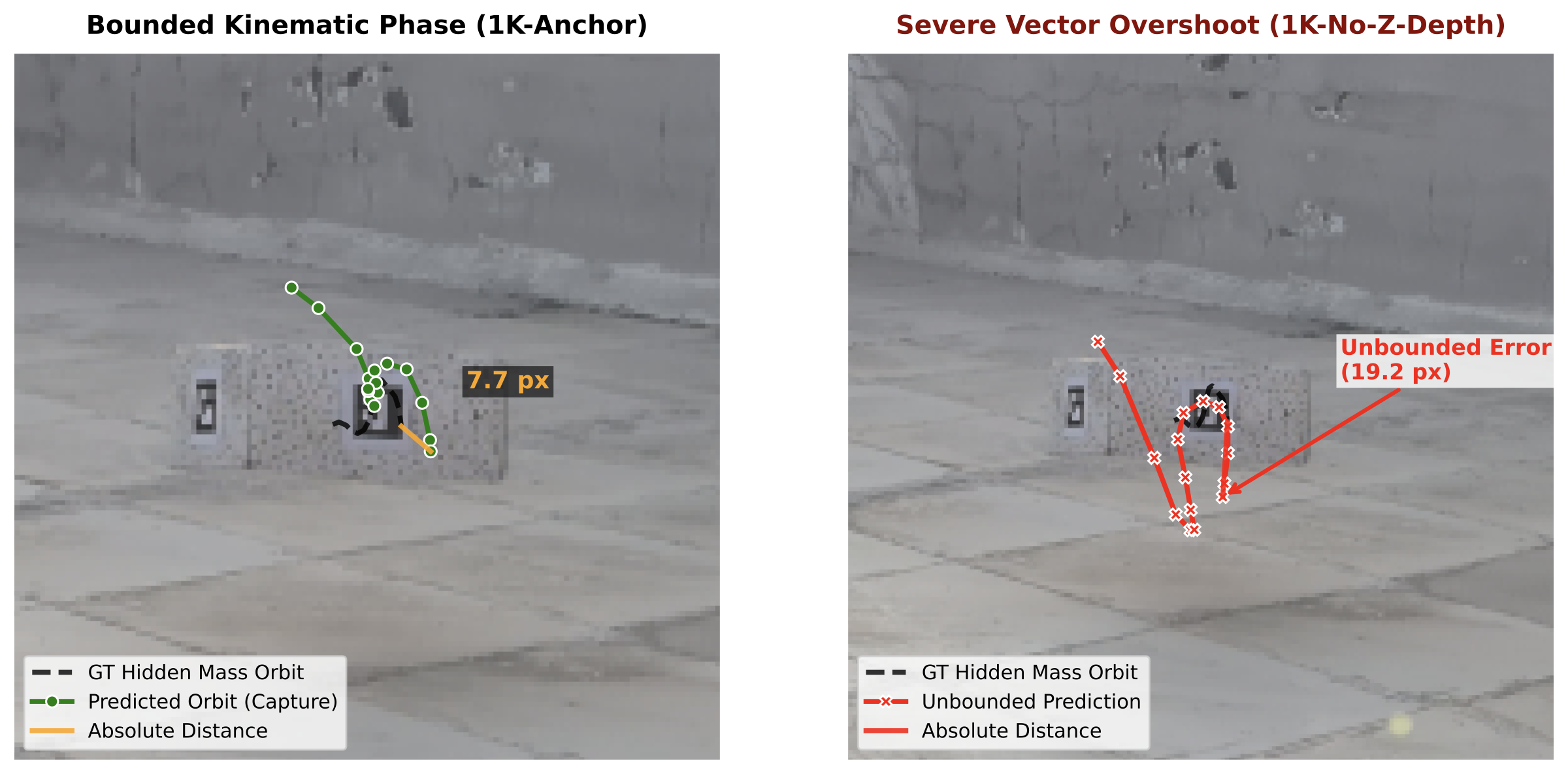}
  \caption{\textbf{The Vector Overshoot Artifact.} (Left) An architecturally complete model (1K-Anchor) correctly learns to bound the hidden mass phase within realistic 3D physical constraints. (Right) Ablating the absolute 1D Z-Depth regularizer (1K-No-Z-Depth) causes the 2D spatial extractions to lose physical scale constraints, leading to a severe, unregularized Euclidean explosion far beyond the physical boundaries of the rigid body.}
  \label{fig:vector_overshoot}
\end{figure}

\paragraph{The Overshooting Artifact.}
While under-regularized models (e.g., STATERA-1K-No-Z-Depth) appear to achieve strong Physics Capture ratios (36.21\%), Euclidean analysis reveals this to be a severe overshooting artifact. Lacking the sufficient data variance required to learn bounded physical priors, small-scale models make unbounded, high-variance predictions extending far beyond the object's physical limits, driving their N-CoME error to an unusable 45.11\%. Scaling to 50K acts as a strictly necessary physical regularizer; it successfully grounds the predictions to realistic physical boundaries, significantly reducing spatial error while achieving a superior 40.96\% Capture. A side-by-side visual comparison of a bounded kinematic trajectory versus an unbounded vector overshoot failure is provided in Figure \ref{fig:vector_overshoot}.

\paragraph{Classical Solver Rank-Deficiency and ArUco Markers.}
To provide the most rigorous comparison possible, we evaluated point-tracking baselines using the idealized dynamic surface centroid proxy (detailed previously in Table \ref{tab:real}). Surprisingly, this proxy allows modern trackers (e.g., TAPIR) to achieve an 8.37\% Capture Ratio. We identify this phenomenon as \textit{kinematic survivorship bias}. During multi-axis rotation, surface points on the object's extremities experience high angular velocity and severe self-occlusion, causing premature track death. Conversely, points near the true rotational axis (the Center of Mass) experience minimal translation and survive longer. Consequently, the spatial average of the surviving tracks artificially drifts toward the CoM. While this occlusion artifact yields an illusion of kinematic tracking, STATERA's end-to-end latent extraction maintains absolute empirical superiority across all metrics (40.96\% Capture, albeit with an overshooting Euclidean penalty), suggesting that internal mass estimation requires holistic temporal priors rather than relying on surface tracking artifacts. Crucially, the ArUco markers \cite{aruco2014} serve only as extrinsic anchors for ground-truth pose estimation and are never provided as explicit state inputs. While they are visibly present in the RGB input as high-contrast textural gradients, the frozen V-JEPA backbone was self-supervised on millions of natural, unconstrained videos devoid of such markers. Its latent space is inductively biased toward natural motion primitives rather than artificial textural heuristics, allowing STATERA to isolate the underlying kinematics despite these textural distractors.

\section{Benchmark Profile Visualization}

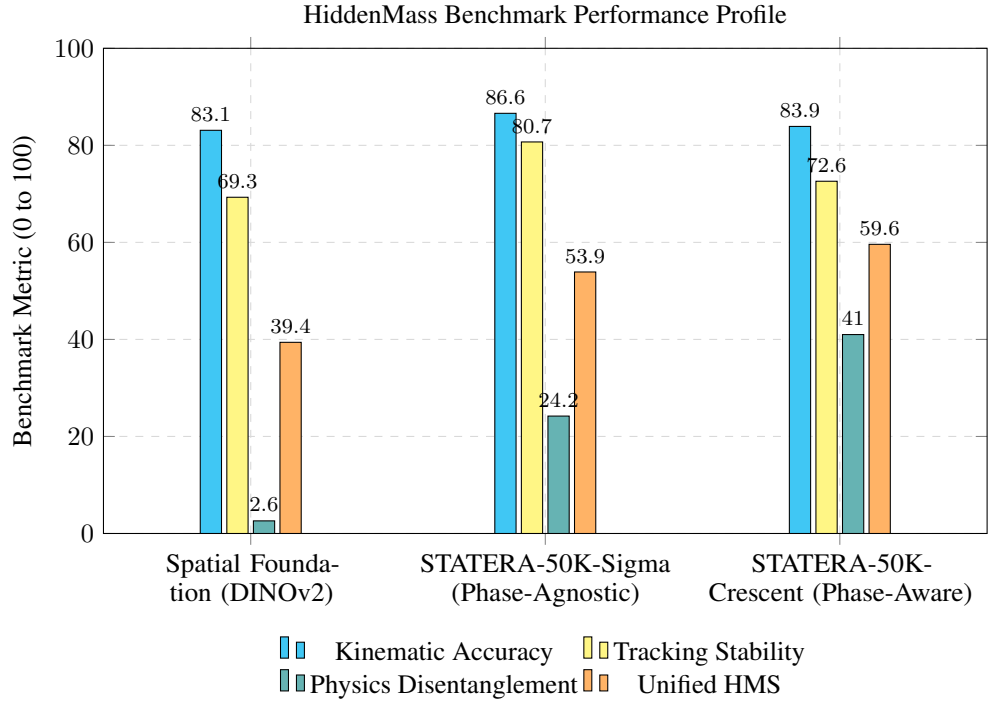
\begin{figure}[ht]
\centering
\begin{tikzpicture}
\begin{axis}[
    ybar,
    bar width=8pt,
    width=0.95\textwidth,
    height=8cm,
    enlarge x limits=0.25,
    legend style={at={(0.5,-0.20)}, anchor=north, legend columns=2, draw=none},
    ylabel={Benchmark Metric (0 to 100)},
    symbolic x coords={Spatial Foundation (DINOv2), STATERA-50K-Sigma (Phase-Agnostic), STATERA-50K-Crescent (Phase-Aware)},
    xtick=data,
    x tick label style={text width=3.5cm, align=center},
    nodes near coords,
    nodes near coords align={vertical},
    nodes near coords style={font=\footnotesize},
    ymin=0, ymax=100,
    grid=major,
    grid style={dashed, gray!30},
    title={HiddenMass Benchmark Performance Profile}
]

\addplot[fill=cyan!60] coordinates {(Spatial Foundation (DINOv2), 83.1) (STATERA-50K-Sigma (Phase-Agnostic), 86.6) (STATERA-50K-Crescent (Phase-Aware), 83.9)};
\addplot[fill=yellow!60] coordinates {(Spatial Foundation (DINOv2), 69.3) (STATERA-50K-Sigma (Phase-Agnostic), 80.7) (STATERA-50K-Crescent (Phase-Aware), 72.6)};
\addplot[fill=teal!60] coordinates {(Spatial Foundation (DINOv2), 2.6) (STATERA-50K-Sigma (Phase-Agnostic), 24.2) (STATERA-50K-Crescent (Phase-Aware), 41.0)};
\addplot[fill=orange!60] coordinates {(Spatial Foundation (DINOv2), 39.4) (STATERA-50K-Sigma (Phase-Agnostic), 53.9) (STATERA-50K-Crescent (Phase-Aware), 59.6)};

\legend{Kinematic Accuracy, Tracking Stability, Physics Disentanglement, Unified HMS}
\end{axis}
\end{tikzpicture}
\caption{\textbf{HiddenMass Benchmark Profile.} The Unified HiddenMass Score (HMS, orange) is calculated as a weighted composite: 25\% Kinematic Accuracy, 25\% Tracking Stability, and 50\% Physics Disentanglement, explicitly penalizing statistical reward-hacking. Low Disentanglement (teal) reveals that DINOv2 and STATERA-50K-Sigma are deceived by visual and heuristic aliasing, defaulting closer to the geometric centroid. STATERA-50K-Crescent actively breaks this heuristic, achieving substantial positive physical disentanglement and securing the highest overall HMS (59.6).}
\label{fig:benchmark_profile}
\end{figure}

This section provides a unified graphical representation of the model evaluation. As shown in Figure \ref{fig:benchmark_profile}, the performance profile explicitly visualizes the trade-offs, prioritizing true physical disentanglement over statistically safe, high-entropy approximations.

\section{Broader Impacts}

The successful integration of pure kinematic tracking from calibrated monocular video poses significant implications for embodied AI and automation systems. By enabling robotic systems to deduce internal mass properties without explicit sensor instrumentation or manual calibration, STATERA reduces the barrier to entry for robust autonomous manipulation in unstructured environments. However, the potential for dual-use applications exists, particularly in contexts where real-time tracking of hidden payloads could be leveraged for unverified surveillance or defense applications. We emphasize that this architecture is intended to enhance safety and predictability in assistive robotics, and any deployment should be subject to rigorous ethical review regarding privacy and intent.

\end{document}